\documentclass[pdflatex,sn-mathphys-ay]{sn-jnl}
\usepackage{graphicx}
\usepackage{multirow}
\usepackage{amsmath,amssymb,amsfonts}
\usepackage{booktabs}
\usepackage{textcomp}
\usepackage{xcolor}
\newcommand{\orcidlink}[1]{\href{https://orcid.org/#1}{\raisebox{-0.2ex}{\includegraphics[height=2.1ex]{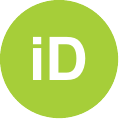}}}}
\makeatletter
\renewcommand\normalsize{\@setfontsize\normalsize{11}{14}%
  \abovedisplayskip 11\p@ \@plus3\p@ \@minus6\p@
  \abovedisplayshortskip \z@ \@plus3\p@
  \belowdisplayshortskip 6.5\p@ \@plus3.5\p@ \@minus3\p@
  \belowdisplayskip \abovedisplayskip
  \let\@listi\@listI}
\renewcommand\small{\@setfontsize\small{10}{12}}
\renewcommand\footnotesize{\@setfontsize\footnotesize{9}{11}}
\renewcommand\scriptsize{\@setfontsize\scriptsize{8}{9.5}}
\def\tablecaptionfont{\reset@font\fontsize{10}{12}\selectfont}
\def\tablebodyfont{\reset@font\fontsize{10}{12}\selectfont}
\def\tablecolheadfont{\reset@font\fontsize{10}{12}\selectfont\bfseries\boldmath}
\def\tablefootnotefont{\reset@font\fontsize{9}{11}\selectfont}
\def\figurecaptionfont{\reset@font\fontfamily{\rmdefault}\fontsize{10}{12}\selectfont}

\def\abstractheadfont{\reset@font\fontsize{11}{14}\bfseries\selectfont\titraggedcenter}
\def\abstractsubheadfont{\reset@font\fontsize{11}{14}\bfseries\selectfont}
\def\abstractfont{\reset@font\fontsize{11}{14}\selectfont\leftskip=0pt\rightskip=0pt\parfillskip=0pt plus 1fil}
\def\keywordfont{\reset@font\fontsize{11}{14}\selectfont\leftskip=0pt\rightskip=0pt}
\makeatother
\normalsize

\begin{document}
\title[RPA vs. Autonomous Agents in Healthcare]{Performance, Scalability and Maintenance: An Experimental Study of RPA vs. Autonomous Agents in the Healthcare Domain}
\author*[1]{\fnm{Daniel} \sur{Schulz\,\orcidlink{0009-0006-7266-0552}}}\email{danielschulz@mx2.unisc.br}
\author[1]{\fnm{Pedro Cabral} \sur{Buchaim\,\orcidlink{0009-0009-7595-0020}}}
\author[1]{\fnm{Bernardo} \sur{Soares\,\orcidlink{0009-0003-8627-5354}}}
\author[1]{\fnm{Fernando Rafael Ensslin} \sur{Schwengber\,\orcidlink{0009-0009-9829-7271}}}
\author[1]{\fnm{Lorenzo} \sur{Farias\,\orcidlink{0009-0000-2219-1889}}}
\author[2]{\fnm{Daniela Duarte} \sur{da Silva Bagatini\,\orcidlink{0000-0001-8548-5841}}}
\author[3]{\fnm{Ana Paula Wernz da Cunha} \sur{M\"uller\,\orcidlink{0000-0002-1919-7152}}}

\affil*[1]{\orgdiv{Department of Engineering, Computer Science and Architecture}, \orgname{University of Santa Cruz do Sul (UNISC)}, \orgaddress{\city{Santa Cruz do Sul}, \state{Rio Grande do Sul}, \country{Brazil}}}
\affil[2]{\orgdiv{Graduate Program in Industrial Systems and Processes (PPGSPI)}, \orgname{University of Santa Cruz do Sul (UNISC)}, \orgaddress{\city{Santa Cruz do Sul}, \state{Rio Grande do Sul}, \country{Brazil}}}
\affil[3]{\orgname{Cl\'inica de Mastologia Dra. Ana Paula Muller Ltda.}, \orgaddress{\city{Porto Alegre}, \state{Rio Grande do Sul}, \country{Brazil}}}

\abstract{Healthcare information systems are fragmented, and retrieving data from legacy web portals is a recurrent need. Robotic Process Automation (RPA) is deterministic but sensitive to interface changes, whereas autonomous agents based on large language models (Agent Process Automation, APA) promise adaptability at a higher computational cost. Following Design Science Research, we developed and compared an RPA solution (Playwright) and an APA solution (browser-use, LangGraph and Gemini 2.5 Flash-lite) to retrieve mammography and breast ultrasound reports from a laboratory portal, treating reports solely as informational artifacts. Across two rounds (T and T+7), RPA reached 98.90\% accuracy and 98.27\% F1-score versus 96.49\% and 94.48\% for APA, was about 13 times faster (16.07 s vs. 207.21 s per patient) and cost about six times less per report (USD 0.030 vs. 0.187). Both reached 100\% step success rates; APA varied more over time. For a stable interface, RPA was superior.}
\maketitle

\section{Introduction}

Process automation has been used to achieve results with minimal human intervention. Robotic Process Automation (RPA) reduces costs and execution time and allows routine tasks to be performed with lower error rates \citep{ribeiro2023}. However, RPA is fragile in dynamic environments such as the web: periodic changes in page layouts require frequent reconfiguration, and the need to adapt to each website reveals its low capacity for generalization \citep{posedaru2024}. The emergence of Large Language Models (LLMs) enabled LLM-based autonomous agents, which are being explored as an alternative to traditional RPA and are here referred to as Agent Process Automation (APA). Recent studies indicate greater adaptability of agents to dynamic interfaces \citep{prucha2025}, but also higher computational demand and operational cost, especially in large-scale tasks \citep{huangjoung2026}.

In healthcare, the fragmentation of information systems hampers data integration and access. Retrieving laboratory reports from separate web portals is a typical case: the task is repetitive, the interfaces are legacy and non-standardized, and the data are sensitive. Although AI and RPA have been associated with operational efficiency and traceability in healthcare when embedded in adequate governance and security architectures \citep{ferber2025,nimkar2024,park2025}, the comparison between RPA and agent-based automation on real healthcare processes remains unexplored (Section 2).

This article presents the development and comparative analysis of RPA and APA solutions for automating the retrieval, extraction and consolidation of mammography and breast ultrasound reports, treating the reports exclusively as informational artifacts, without any automated clinical inference. Following Design Science Research (DSR) \citep{hevner2004,peffers2007}, both solutions were implemented as artifacts and evaluated in a real case study provided by the ClickPalm platform, a Personal Health Record platform that centralizes and integrates patients' clinical information. The evaluation relies on predefined technical indicators: classification performance (accuracy, precision, recall and F1-score), operational performance (execution time), cost, success rate per process step and temporal stability. The contributions are (i) quantitative evidence on the trade-off between computational cost and flexibility of RPA and APA in a real, sensitive healthcare use case, and (ii) a reference for architectural decisions in similar clinical automation projects.

The article is organized as follows. Section 2 presents the background and the research gap; Section 3 describes the method and experimental procedures; Section 4 details the development of the RPA and APA solutions; Section 5 presents the results; Section 6 discusses theoretical and practical implications and limitations; and Section 7 concludes. The declarations, including ethics approval, follow the conclusion.

\section{Background}

\subsection{RPA and agent-based automation}

Automation has been used to obtain results with minimal human intervention. According to \citet{ribeiro2023}, RPA, besides reducing costs and execution time, allows routine tasks to be performed with a lower error rate. Although frequently associated with financial efficiency and high return on investment, the literature also points to limitations in dynamic environments such as the web, where periodic updates of page layout demand frequent reconfiguration. Moreover, the need to adapt to different websites evidences its low generalization capacity \citep{posedaru2024}.

The creation of LLMs enabled LLM-based autonomous agents, which have been explored as an alternative to traditional RPA. \citet{prucha2025} indicate that autonomous agents show greater adaptability to dynamic interfaces, especially in scenarios requiring frequent reconfiguration of the interaction environment and contextual decision-making. Despite these advantages, \citet{huangjoung2026} mention that autonomous agents may demand greater computational capacity and higher operational costs, especially in large-scale tasks.

\subsection{Automation in healthcare information systems}

In healthcare, the fragmentation of information systems is a recurrent challenge, hindering the integration and efficient access to data. Studies by \citet{nimkar2024}, \citet{park2025} and \citet{ferber2025} show that solutions based on Artificial Intelligence and RPA can contribute to operational efficiency, support administrative activities, improve process traceability and support decision-making, provided they are embedded in adequate information governance and security architectures.

\subsection{LLM-based web data extraction}

Recent architectures have explored LLMs and their variants for information extraction in web environments. \citet{huangjoung2026} investigated Multimodal Large Language Models (MLLMs) for extracting information from websites using the WebScraper framework, which specializes the generalist GUI agent \emph{Computer Use} as an information extractor through specific tools (Parse Tool and Merge Tool) and structured prompts. Tested on six news websites and on platforms such as Amazon and Momo, the framework showed significant accuracy improvement over generic agents.

Previously, \citet{huang2024} proposed AutoScraper, a framework that uses the DOM (Document Object Model) hierarchy of HTML pages together with LLMs to generate XPath selectors. Based on top-down and step-back operations, the model navigates the DOM tree in search of the nodes that contain the desired information and then builds XPath expressions capable of extracting their contents automatically. AutoScraper was tested on the SWDE datasets (pages from 80 sites across eight domains), Extended SWDE (a version of SWDE with more detailed annotation of 21 sites) and DS1 (30 large real sites), all with manual annotation of the information to be extracted, generally outperforming baseline approaches such as Chain of Thought and Reflexion. \citet{posedaru2024} developed an approach for automated data extraction from web pages using language models, such as ChatGPT, to identify CSS selectors and generate NodeJS scraping functions. Once generated, these functions can be run locally, reducing AI token consumption. Tests on three websites yielded satisfactory results.

\subsection{Research gap}

Although these studies show growing interest in using LLMs for automation in mutable environments, their tests were conducted in generic settings, such as news portals, e-commerce platforms and sets of pages built for experimentation. There is also an absence of comparative experimental studies that systematically evaluate the performance of RPA and autonomous agents in real healthcare processes, especially those involving heterogeneous systems and interfaces. This article addresses this gap.

\section{Research method}

This applied research adopts a quantitative, experimental approach based on Design Science Research (DSR; \citealp{hevner2004,peffers2007}). To evaluate the systems developed over the iterative cycles, we used classification metrics (accuracy, precision, recall and F1-score), operational performance and cost metrics, success rate per workflow step, and temporal stability. The study is structured in four methodological stages, detailed below.

\subsection{Methodological procedures}

\textbf{Stage 1 -- Problem awareness.} A literature review was conducted in Scopus, IEEE, Web of Science and arXiv. It identified a gap: the absence of comparative experimental studies that systematically evaluate RPA and APA in the face of the heterogeneity of healthcare systems. From this gap, the scope of the study was delimited, motivated by the need to understand the technical trade-offs between these approaches in real clinical environments.

\textbf{Stage 2 -- Solution proposal.} This stage consisted of planning and conceptually designing RPA- and APA-based solutions to extract information from mammography and breast ultrasound reports, meeting the requirements of a real case study provided by the ClickPalm platform.

\textbf{Stage 3 -- Development.} First, the execution flow that the RPA- and APA-based robots should follow was defined, followed by the steps needed to execute it. A control table with the eligible patients was also required (Section 3.5). As shown in Fig. 1, the execution flow was defined in nine steps. The robot reads the table and identifies patients with pending exam downloads (1); the patient names are organized in a processing queue accessible to the robot (2), which performs a single login to the predefined laboratory portal (3). After authentication, the robot locates the patient by name (4) and selects the mammography and breast ultrasound exams performed from 2024 to 2025 (5). It then downloads the respective reports in PDF format, excluding the pre-operative localization letters, whose names are similar to those of the target exams but which are not the documents of interest (6). The download history is then recorded (7). If there is another patient to be processed, the flow returns to the name search step; otherwise, it proceeds to step 9 (8). Finally, the spreadsheet status is updated, indicating that the patients' exams were downloaded (9). With the conclusion of this flow, the development stage was considered complete.

\begin{figure}[!ht]
\centering
\includegraphics[width=\textwidth,height=0.82\textheight,keepaspectratio]{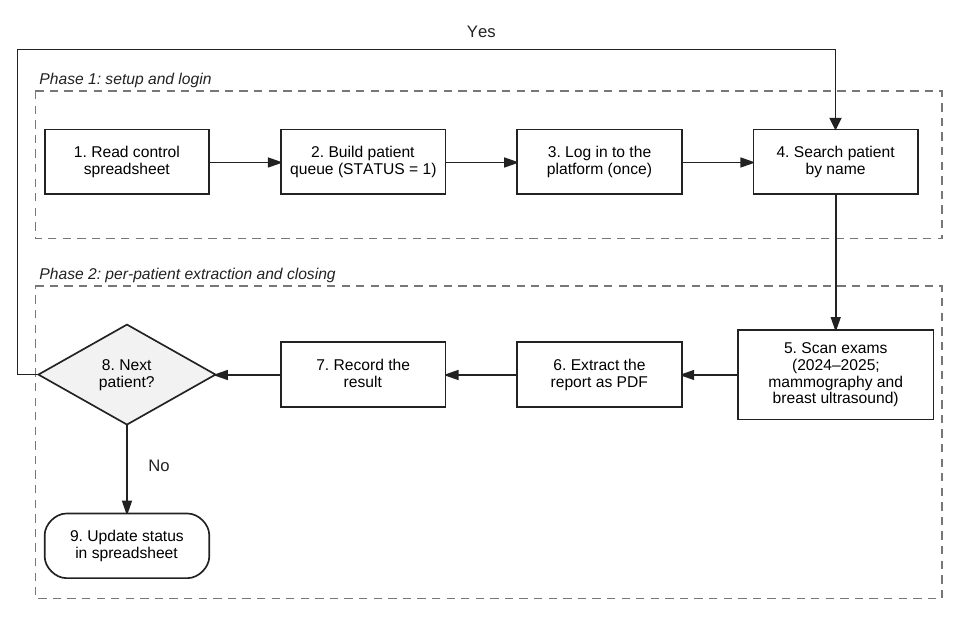}
\caption{Macro process flowchart and business rules for report extraction}\label{fig:1}
\end{figure}

\textbf{Stage 4 -- Evaluation.} A quantitative evaluation of the RPA- and APA-based solutions was carried out, considering metrics related to exam collection performance, operational performance, cost, success rate in each workflow step and temporal stability. Detailed in Section 3.6, these metrics allowed an objective comparison of the two solutions under experimental conditions and established the criteria used to analyze the results presented in Section 5.

\subsection{Case study}

This study used the ClickPalm platform as the experimental environment, in which controlled and comparable scenarios were implemented for the two automation approaches. Within this work, the platform provided a real case study on the automated extraction of mammography and breast ultrasound reports of patients by a mastologist physician.

Before the automation solutions were implemented, the reports were collected manually by the responsible physician. For each patient, it was necessary to locate the name in the patient list, log in to the laboratory portal, locate the corresponding exam and manually download the report in PDF format. The ClickPalm case study was adopted as the experimental scenario because it represents an automation process in a healthcare environment, characterized by legacy systems, non-standardized interfaces and the need for accuracy in the extraction of sensitive data. These characteristics allowed the RPA- and APA-based solutions to be evaluated under conditions representative of the collection process performed in the clinical context. The study thus enabled a technical comparison of the two approaches and an analysis of their performance in a real application context, considering aspects related to operational efficiency and execution of the collection process.

\subsection{Experimental environment}

The ClickPalm platform and the laboratory portal are distinct systems with different functions in this study. While ClickPalm provided the context of the case study, the reports were extracted directly from the portal of the laboratory responsible for issuing them, since there was no integration between the two systems. The website used for the tests is therefore the laboratory service portal, where the exams of the patients included in the study were available. With the authorization of the responsible physician, the solutions used her access credentials to the portal, and collection was limited to the minimum data necessary for the study's objectives, in observance of the necessity principle of the Brazilian General Data Protection Law (LGPD; \citealp{brazil2018}), as detailed in the Declarations.

The website server showed no significant performance variation between different shifts. During development, tests were run in the morning, afternoon and evening, with no relevant change in performance. At specific moments, slowness was observed when loading the exam list of some patients, which normalized shortly afterwards. Overall, the observed results indicate server stability during the testing period.

The main problem identified at the beginning of the tests was related to the website interface: there was no specific button to generate and download the report directly as a PDF, and the print option had to be used to generate the file. In addition, each exam in the list was identified by a card with distinct titles, and this lack of standardization hinders the identification of duplicate exams and the distinction between mammography and breast ultrasound reports and pre-operative localization letters, which are unrelated to the exams of interest. While exploring the portal, it was found possible to pin, beside the exam, a bar that displayed the content of the report directly in the website interface. From this discovery, the solutions began to identify the report content without having to interpret the non-standardized cards, which in some cases displayed no information at all, even for breast ultrasound reports. It was thus possible to work around the limitations imposed by the portal interface and complete the development of the solutions.

\subsection{Execution infrastructure}

To ensure equivalent experimental conditions for comparing the metrics of the RPA- and APA-based solutions, both were run in the AWS cloud computing environment using containers. The use of containers is justified by the possibility of reducing infrastructure costs compared with environments that require one virtual machine per application, since containers share the same operating system and computational resources of the host. This approach reduces the memory and processing overhead associated with virtualization and allows greater application density per host. Where applicable, it can also reduce costs related to operating system licensing and other infrastructure components.

Another relevant advantage is scalability, since containers can be started and stopped quickly, allowing computational resources to be allocated according to demand. The RPA- and APA-based solutions are compatible with this execution model because they are processes that can be isolated in containers, provided the specific performance, security and configuration requirements of each solution are observed. Running both solutions in containers in the cloud thus provides greater flexibility in resource allocation and facilitates the management of the computing environment.

\subsection{Data}

The dataset comprised mammography and breast ultrasound exams of the responsible physician's patients, made available through the laboratory platform, with prior authorization of the patients for research use, formalized by an Informed Consent Form (TCLE; see Declarations). Only exams from 2024 to 2025 were considered, since earlier exams were not relevant to the scope of the study.

During the experimental period, 69 patients were processed, corresponding to 71 exams. Records identified as pre-operative localization letters were excluded from the analysis because, although their names are similar to those of the exams of interest, they were not mammography or breast ultrasound reports.

Because of the limited number of API keys available for running the APA-based solution, whose usage quotas were renewed daily, patient processing was distributed in batches of up to 16 patients per day over five consecutive days, the last batch comprising five patients. All processing was run in two rounds, at T and at T+7, with the same patients, batches and reference criteria. This strategy circumvented the daily quota limit of LLM API calls while maintaining comparability between the solutions, since both processed exactly the same batches of patients on the same dates.

\subsection{Evaluation metrics}

To systematically compare the performance of the RPA and APA solutions, four categories of evaluation metrics were defined.

\bmhead{Exam collection performance}

To evaluate exam collection performance, each execution was analyzed through a confusion matrix, considering:

\begin{itemize}
\item
  \textbf{True Positive (TP):} authorized, non-duplicate exam correctly collected;
\item
  \textbf{False Positive (FP):} unauthorized exam collected, or authorized exam collected in duplicate (e.g., a pre-operative localization letter downloaded by mistake);
\item
  \textbf{False Negative (FN):} authorized, non-duplicate exam that should have been collected but was not;
\item
  \textbf{True Negative (TN):} unauthorized exam, or already-collected authorized exam, that was not collected again.
\end{itemize}

Each round (T and T+7) was evaluated separately, and the matrices presented in Section 5.1 correspond to the sum of the two rounds. Each decision corresponds to one exam listed in the portal for the processed patients, classified as authorized (mammography or breast ultrasound report from 2024 to 2025, non-duplicate) or unauthorized (other exams, localization letters and duplicates). From these values, the following evaluation metrics were calculated:

\begin{itemize}
\item
  \textbf{Accuracy:} proportion of correct system decisions, considering exams correctly collected and correctly not collected;
\end{itemize}

\[
\mathrm{Accuracy}=\frac{TP+TN}{TP+TN+FP+FN}\in[0,1]
\]

\begin{itemize}
\item
  \textbf{Precision:} proportion of correctly collected exams among all exams collected by the system, indicating the ability to avoid undue collection;
\end{itemize}

\[
\mathrm{Precision}=\frac{TP}{TP+FP}\in[0,1]
\]

\begin{itemize}
\item
  \textbf{Recall:} proportion of correctly collected exams among all exams that should have been collected, indicating the ability to identify eligible exams;
\end{itemize}

\[
\mathrm{Recall}=\frac{TP}{TP+FN}\in[0,1]
\]

\begin{itemize}
\item
  \textbf{F1-score:} harmonic mean of Precision and Recall, summarizing system performance on these two dimensions.
\end{itemize}

\[
F1=\frac{2\times\mathrm{Precision}\times\mathrm{Recall}}{\mathrm{Precision}+\mathrm{Recall}}\in[0,1]
\]

Because of duplicate exams in the source portal, both RPA and APA downloaded both versions. To handle this, each solution independently implemented an automatic duplicate-checking module based on textual similarity between the downloaded documents. Documents with similarity equal to or greater than 90\% were considered by the respective solution to refer to the same report and counted as a single occurrence in the confusion matrix. This procedure increased the accuracy obtained by both solutions.

\bmhead{Operational performance and cost}

For each solution, the following operational performance and cost metrics were recorded. The report, rather than the batch, was adopted as the reference unit, given the small number of executions (five batches) relative to the volume of reports processed (more than 140 per solution across the two rounds), which provided a more robust cost estimate.

\begin{itemize}
\item
  \textbf{Mean execution time per patient:} mean time, in seconds, devoted to processing each patient, computed as the ratio between the total execution time and the number of patients processed;
\end{itemize}

\[
\mathrm{Mean\ time\ (s)}=\frac{\text{total time of all executions}}{\text{number of patients}}
\]

\begin{itemize}
\item
  \textbf{Cost per report:} mean cost of obtaining a single report, considering infrastructure costs and, for APA, also LLM API call costs. It is computed as the ratio between the total infrastructure cost (plus the LLM API cost, when applicable) and the number of reports successfully downloaded. For APA, the LLM API cost is computed from the input/output tokens of each call, multiplied by the per-token price of the model used (Gemini 2.5 Flash-lite).
\end{itemize}

\[
\mathrm{Cost\ per\ report\ (USD)}=\frac{\text{infrastructure cost (+ LLM API cost, for APA)}}{\text{exams successfully downloaded}}
\]

\bmhead{Success rate per step}

To identify the workflow steps in which each solution is more or less reliable, performance was evaluated individually for each step, considering:

\begin{itemize}
\item
  \textbf{Login rate:} proportion of authentication attempts completed successfully;
\end{itemize}

\[
\mathrm{Login\ rate\ (\%)}=\frac{\text{successful logins}}{\text{successful logins}+\text{unsuccessful logins}}\times100
\]

\begin{itemize}
\item
  \textbf{Identification/search rate:} proportion of registered patients correctly located in the portal after authentication;
\end{itemize}

\[
\mathrm{Search\ rate\ (\%)}=\frac{\text{successful searches}}{\text{registered patients}}\times100
\]

\begin{itemize}
\item
  \textbf{Complete download rate:} proportion of registered patients for whom all expected exams were successfully collected. In both rates, the denominator is the 65 patients with an active registration in the laboratory; the four patients with no registration were excluded (Section 6.1).
\end{itemize}

\[
\mathrm{Complete\ download\ rate\ (\%)}=\frac{\text{patients with all exams downloaded}}{\text{registered patients}}\times100
\]

\bmhead{Temporal stability}

To evaluate the consistency of the solutions over time, we adopted the comparative measurement protocol between days T and T+7, as proposed by \citet{huangjoung2026}. All patients were reprocessed at two distinct moments, first at T and again seven days later at T+7, keeping the same set of patients and the same reference criteria (gold standard). At each moment, the classification metrics described in Section 3.6.1 (Accuracy, Precision, Recall and F1-score) were calculated. Temporal stability was assessed by the absolute difference between the values of each metric obtained at T and T+7. Smaller variations indicate greater stability of the solution in the face of possible changes in the portal, such as layout updates and exam availability, or variations in data conditions over time.

\section{Solution development}

This section describes the development of the two compared solutions. Section 4.1 presents the RPA, implemented with Playwright, and Section 4.2 presents the APA, based on browser-use and LangGraph. Both follow the flow defined in Fig. 1 and apply the same business criteria, differing only in the way they are executed.

\subsection{Traditional RPA}

The RPA solution was developed in Python using the Playwright library (version 1.60.0) for browser automation; its source code is available as Online Resource 2. Playwright was chosen for its deterministic execution model and its fine control over the loading of pages, frames and dynamic elements, characteristics essential for dealing with a legacy web portal whose interface was not designed for automation.

The overall architecture and the processing flow of the RPA are shown in Fig. 2. The solution was structured in modules with well-defined responsibilities: \emph{main.py}, which orchestrates execution, including reading the patient queue, opening the browser, authentication and iteration over the patients; \emph{core.py}, which concentrates the portal interaction logic, including patient search, mapping and exam download; \emph{data\_manager.py}, responsible for managing the download history and normalizing names; \emph{report\_manager.py}, responsible for generating execution reports, error logging and telemetry; and \emph{config.py}, which centralizes configuration parameters, interface selectors and filtering criteria.

A central principle of the architecture is to maintain a single browser session throughout the patient batch: login is performed once and, at the end of each patient's processing, the robot returns to the search screen through an internal application action, without reloading the page or re-authenticating. A full new login is used only as a contingency strategy for cases in which the session is unexpectedly terminated. This approach reduces total execution time and the number of requests to the portal server.

To read and update the patient queue, the gspread library was used together with pandas, allowing direct integration with a Google Sheets spreadsheet. Patients with pending status (STATUS = 1) are read at the start of execution and, at the end of each patient's processing, the status is updated in the spreadsheet itself, signaling completion or the need for reprocessing.

The filtering of relevant exams was implemented through two complementary verification layers. In the first layer, the exam date is checked against the interval defined in the configuration parameters, together with the presence of keywords associated with the exams of interest (mammography, breast ultrasound, among variations) in the text of the card displayed in the listing. In the second layer, after the report is opened, its entire content, including any internal frames, is analyzed for textual markers indicating that it is a pre-operative localization or procedure letter rather than a diagnostic report. This second check was necessary because the name of the card displayed in the interface does not always correspond to the actual content of the document, making filtering based only on the exam title insufficient.

For extracting PDF reports, a chain of strategies was implemented, executed in priority order, each triggered only when the previous one fails: (1) direct download via the file URL, when available in an iframe; (2) extraction of the PDF URL through JavaScript evaluation on the page; (3) extraction of the content from a blob object dynamically generated by the browser; and (4) conversion of the HTML page into PDF, used as a last resort when the portal does not provide the report as a file. This layered architecture was adopted because of the heterogeneity of the portal interface, which presents different report delivery mechanisms depending on the exam type and the moment of consultation.

Download duplication control is performed through a local history file that records a unique identifier for each processed exam, composed of the patient's name and identifier, the exam date and the exam name, avoiding reprocessing in future executions. Failures during processing, such as a missing report or PDF extraction errors, are recorded in a structured way, allowing later manual intervention without interrupting the processing of the remaining exams and patients in the queue. Both solutions were instrumented to record, for each processed exam, the data needed to calculate the evaluation metrics described in Section 3.6, enabling later computation of indicators such as exam classification, success rates per step and temporal stability.

During the tests in the cloud environment, failures in the collection of certain reports were observed in part of the executions, associated with the same false positives discussed later. It was first considered that differences in the viewport dimensions of the page in headless mode on AWS were the cause, a hypothesis later discarded. Analysis of the execution logs identified the root cause: the iframe-based extraction strategy, which in some cases selected an incorrect frame by prioritizing the largest available frame, not always the one corresponding to the report, or attempted to access a frame that was still loading at the time of extraction. After the frame selection logic was corrected, these failures were no longer observed in subsequent tests.

Finally, to enable execution in the cloud environment, as described in Section 3.4, the solution was containerized with Docker, with an image based on Playwright for Python, ensuring the availability of the browsers and system dependencies required to run in headless mode, without a graphical interface.

Fig. 2 summarizes the technical flow of the RPA in three phases: configuration and login; per-patient processing, which includes the name search and the selection of valid exams; and per-exam processing, in which duplication and report validity are checked and the PDF is extracted through the chain of four strategies (iframe, JavaScript, blob and HTML to PDF).

\begin{figure}[!ht]
\centering
\includegraphics[width=\textwidth,height=0.82\textheight,keepaspectratio]{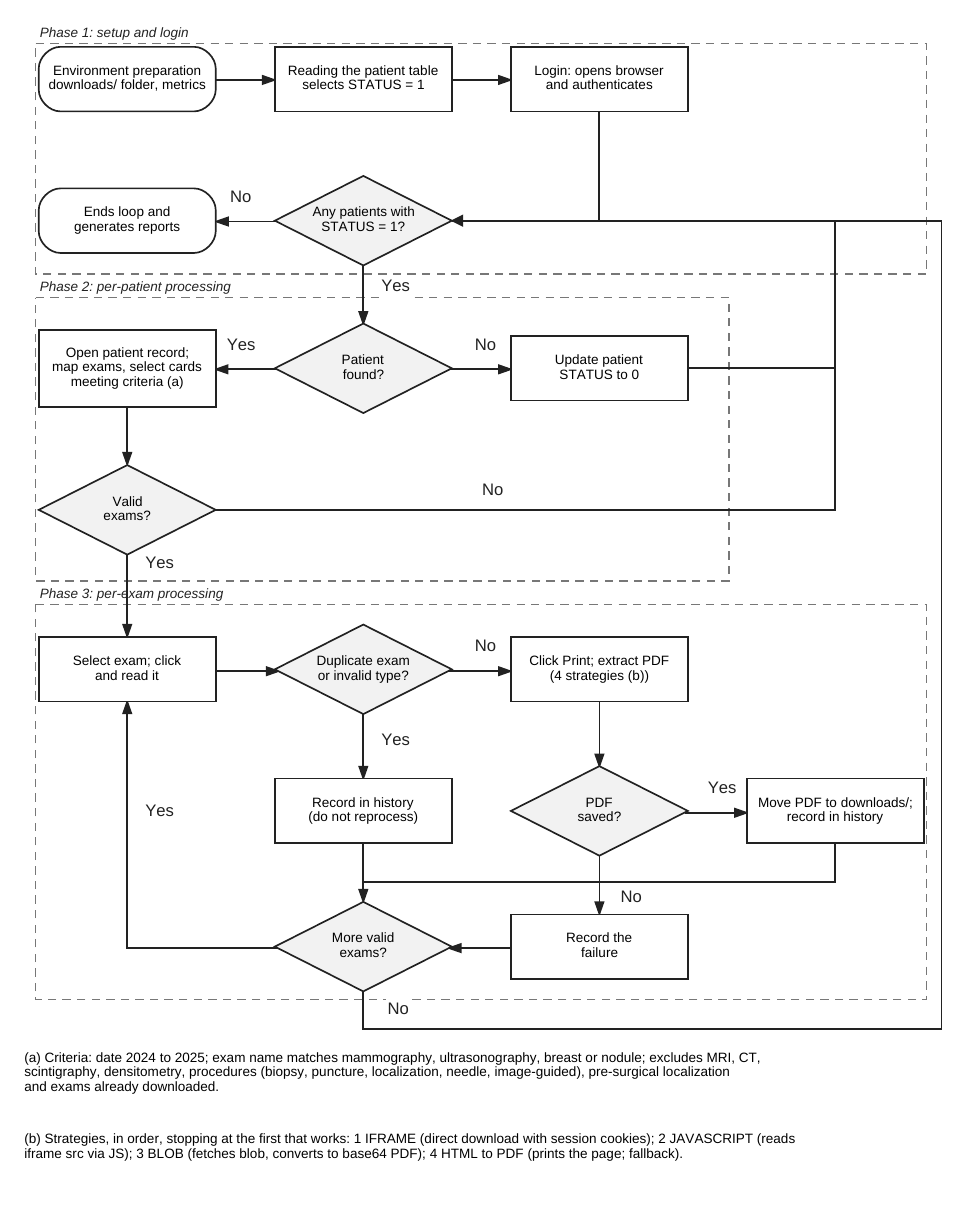}
\caption{Technical flowchart of the RPA solution (Playwright)}\label{fig:2}
\end{figure}

\subsection{APA (Agent Process Automation)}

The APA solution (source code in Online Resource 1; modified version of the browser-use library in Online Resource 3) was developed in Python using the browser-use library for the autonomous control of the browser by a language-model-based agent, and LangGraph for orchestrating the batch execution flow. Unlike RPA, in which each interaction with the page is explicitly programmed, in APA the agent receives a task described in natural language and determines, at each step, the actions to be performed on the interface, such as navigating, clicking and filling fields, based on its interpretation of the page content.

The overall architecture and the processing flow of the APA are represented in Fig. 3. The project architecture was organized in three main layers: an orchestration layer, composed of \emph{graph.py} and \emph{main.py}, which defines the high-level batch processing flow through a LangGraph state graph (reading the queue → opening the browser → processing each patient, in a loop → finalization); an agent layer, formed by \emph{runner.py}, \emph{prompts.py}, \emph{tools.py} and \emph{extraction.py}, which instantiates the language model, builds the task (prompt) for each patient and provides the tools for executing the actions; and a domain layer, with \emph{filters.py}, \emph{history.py} and \emph{models.py}, which shares with the RPA the same deterministic logic for filtering exams by date and keywords, as well as the download history control. This structure ensures that both solutions apply exactly the same business criteria, differing only in the way they are executed.

Given the sensitive nature of the data handled, which are reports of real patients' exams, the native anonymous telemetry of the browser-use library, which by default sends the library maintainer information such as task success, its description, the number of steps executed and location data, was disabled in the solution configuration. This measure prevented the sending of execution metadata to third-party servers, in line with the principles of protection of sensitive personal data under the LGPD \citep{brazil2018}.

As in the RPA, the APA maintains a single browser session throughout the batch (parameter \emph{keep\_alive=True}), avoiding unnecessary re-authentication between patients. For each patient, the agent receives a dynamically built task (prompt) informing the patient's name, the access credentials and a set of sequential instructions: log in only if necessary; locate the patient, trying first the full name and, if that fails, only the first name; map the target exams (based on year and keywords); and, for each exam, open the report and delegate the capture of the report to a deterministic tool. The prompt was written in English, based on the empirical observation of better model performance in that language, and was deliberately kept lean, concentrating the critical steps of filtering, extraction and duplication control in deterministic tools instead of delegating them to the language model.

The structure of the prompt reflects a set of engineering decisions aimed at reducing the agent's ambiguity and token consumption. First, the task instructs the agent to build in advance a single, complete list of the patient's target exams, processing each item exactly once. This decision was motivated by the need to avoid reprocessing or repetition loops, common in agents that decide their own actions at each step. Second, the prompt explicitly delegates PDF capture to the deterministic tool \emph{download\_exam\_report}, forbidding the agent from trying to extract the report on its own or from using the generic structured-data extraction action (\emph{extract}), an instruction justified, according to the project documentation, by the fact that this generic action is slow and consumes a significant share of the token budget available for the execution. Third, the prompt establishes explicit session persistence rules, instructing the agent never to return to the login screen, never to repeat the patient search, and to treat technical interface error messages (e.g., loss of focus between tabs) as situations to be worked around, not as a reason to restart the process. This approach maintains the session stability principle adopted in the RPA, adapted to the context of an autonomous agent. Finally, the prompt defines a mandatory structured output at the end of the task, validated by a data schema, replacing the interpretation of free text and increasing the predictability and reliability of how the orchestration reads the results.

The main tool implemented is the function \emph{download\_exam\_report}, responsible for capturing the PDF report of the currently open exam, without depending on the language model's interpretation at this step. This tool reuses, through the Chrome DevTools Protocol (CDP), required because of the absence of direct access to the Playwright API in the version of browser-use used, the same textual report verification logic employed in the RPA. The page content is analyzed and compared with markers indicating a procedure letter or an unavailable report, avoiding the undue collection of these documents. When the content is considered valid, PDF extraction is performed through two strategies, in priority order: (1) a network request (FETCH) performed inside the browser itself with the active session credentials, and (2) rendering of the page to PDF through the \emph{Page.printToPDF} command, used as an alternative when the first strategy does not return a valid file.

Download duplication control was implemented in two layers: a short-term one, restricted to the current execution of each patient, avoiding the reopening or reprocessing of the same exam within one cycle, and a persistent one across executions, shared with the RPA through the same history file. At the end of each patient's processing, the agent returns a structured output, validated through a Pydantic schema, containing the patient identifier, the number of exams downloaded and the list of exams reported as unavailable, replacing free-text interpretation with a predictable and verifiable output format. As in the RPA, this instrumentation provides the data needed to calculate the evaluation metrics described in Section 3.6.

As the language model, the solution was designed to support both models from the Google Gemini family and Claude models run via AWS Bedrock, configurable through an environment variable. In the experiments conducted in this study, the Gemini 2.5 Flash-lite model was used, run with a low temperature to favor greater determinism in the agent's decisions, together with a fallback model, triggered automatically in case of unavailability or rate limiting of the main model. Finally, to enable execution in the cloud environment, as described in Section 3.4, the solution was likewise containerized, following the same headless execution logic adopted by the RPA.

Fig. 3 presents the technical flow of the APA in four phases: configuration, queue and orchestration; patient processing, with name search by degradation; exam selection and processing, in which the agent calls the \emph{download\_exam\_report} tool and the report is validated via CDP; and patient consolidation, with the status update in the spreadsheet.

\begin{figure}[!ht]
\centering
\includegraphics[width=\textwidth,height=0.82\textheight,keepaspectratio]{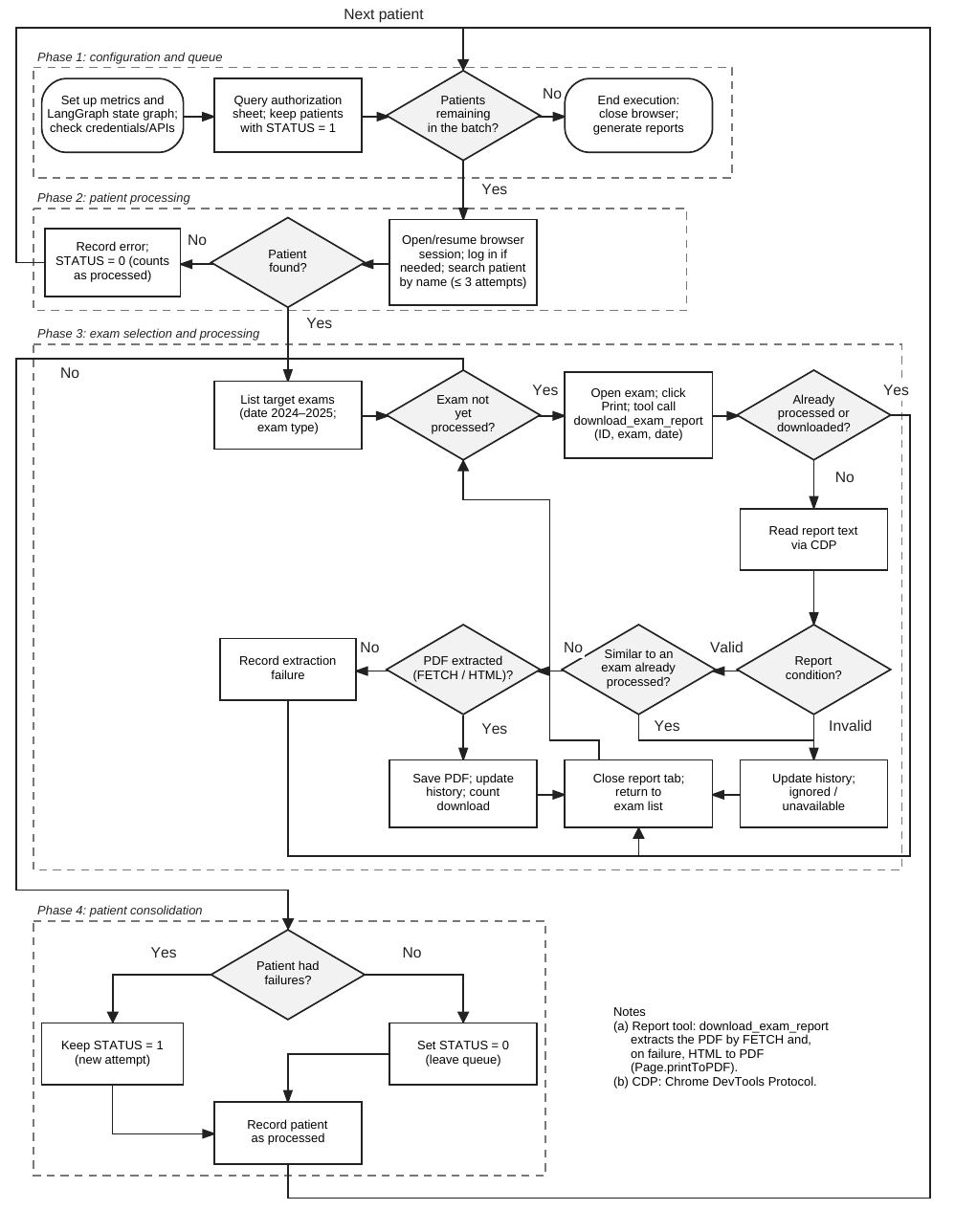}
\caption{Technical flowchart of the APA solution (browser-use and LangGraph)}\label{fig:3}
\end{figure}

\section{Results}

This section presents the results of the comparison between RPA and APA, organized according to the metrics defined in Section 3.6: exam collection performance (5.1), operational performance and cost (5.2), success rate per step (5.3) and temporal stability (5.4). Except in Section 5.4, the values refer to the two collection rounds (T and T+7) summed.

\subsection{Exam collection performance}

Table 1 shows the confusion matrix model adopted: rows indicate the true value (authorized exam or not) and columns the automation's decision (download or not download). Table 2 presents the matrices obtained by the RPA and the APA, and Table 3 the metrics derived from them.

\begin{table}[!ht]
\caption{Confusion matrix model adopted for the classification of the collected exams}\label{tab:1}
\centering
\begin{tabular*}{\textwidth}{@{\extracolsep{\fill}}@{}lcc@{}}
\toprule
True value & Download & Not download \\
\midrule
Authorized & Exam collected (TP) & Exam not collected (FN) \\
Not authorized & Exam collected (FP) & Exam not collected (TN) \\
\bottomrule
\end{tabular*}
\end{table}

\begin{table}[!ht]
\caption{Confusion matrices of the RPA and APA, summing the two executions (T + T+7)}\label{tab:2}
\centering
\begin{tabular*}{\textwidth}{@{\extracolsep{\fill}}@{}lcccc@{}}
\toprule
 & RPA Download & RPA Not download & APA Download & APA Not download \\
\midrule
Authorized & 142 (TP) & 0 (FN) & 137 (TP) & 5 (FN) \\
Not authorized & 5 (FP) & 309 (TN) & 11 (FP) & 303 (TN) \\
\bottomrule
\end{tabular*}
\end{table}

\emph{Total of 456 decisions per solution}

\begin{table}[!ht]
\caption{Classification metrics of exam collection (RPA and APA)}\label{tab:3}
\centering
\begin{tabular*}{0.7\textwidth}{@{\extracolsep{\fill}}@{}lcc@{}}
\toprule
Metric & RPA & APA \\
\midrule
Accuracy & 98.90\% & 96.49\% \\
Precision & 96.60\% & 92.57\% \\
Recall & 100\% & 96.48\% \\
F1-score & 98.27\% & 94.48\% \\
\bottomrule
\end{tabular*}
\end{table}

\subsection{Operational performance and cost}

Table 4 presents the mean execution time per patient and the cost per report of each solution, computed according to the formulas in Section 3.6.2.

\begin{table}[!ht]
\caption{Mean execution time per patient and cost per report}\label{tab:4}
\centering
\begin{tabular*}{0.7\textwidth}{@{\extracolsep{\fill}}@{}lcc@{}}
\toprule
Metric & RPA & APA \\
\midrule
Mean execution time per patient (s) & 16.07 s & 207.21 s \\
Cost per report (USD) & 0.030 & 0.187 \\
\bottomrule
\end{tabular*}
\end{table}

\subsection{Success rate per step}

Table 5 presents, for each solution, the proportion of success in the login, patient identification and exam download steps, according to the formulas in Section 3.6.3.

\begin{table}[!ht]
\caption{Success rate per workflow step}\label{tab:5}
\centering
\begin{tabular*}{0.7\textwidth}{@{\extracolsep{\fill}}@{}lcc@{}}
\toprule
Step & RPA & APA \\
\midrule
Login rate (\%) & 100\% & 100\% \\
Identification/search rate (\%) & 100\% & 100\% \\
Complete download rate (\%) & 100\% & 100\% \\
\bottomrule
\end{tabular*}
\end{table}

\emph{Considering the 65 patients registered in the laboratory; four patients had no active registration (see Section 6.1)}

\subsection{Temporal stability: test–retest T vs. T+7}

Table 6 presents the absolute variation, in percentage points, of each metric between rounds T and T+7. The smaller the variation, the greater the stability.

\begin{table}[!ht]
\caption{Variation of the metrics between T and T+7 (temporal stability)}\label{tab:6}
\centering
\begin{tabular*}{0.7\textwidth}{@{\extracolsep{\fill}}@{}lcc@{}}
\toprule
Metric & RPA & APA \\
\midrule
Accuracy – variation T vs. T+7 (p.p.) & 0.44 & 3.51 \\
Precision – variation T vs. T+7 (p.p.) & 1.31 & 4.26 \\
Recall – variation T vs. T+7 (p.p.) & 0.00 & 7.04 \\
F1-score – variation T vs. T+7 (p.p.) & 0.68 & 5.59 \\
\bottomrule
\end{tabular*}
\end{table}

\section{Discussion}

The results allow a complementary reading of the classification, operational performance and cost metrics of the two solutions.

\subsection{Theoretical implications}

In exam classification, the RPA outperformed the APA in accuracy (98.90\% vs. 96.49\%), precision (96.60\% vs. 92.57\%), recall (100.00\% vs. 96.48\%) and F1-score (98.27\% vs. 94.48\%). The precision difference suggests that the APA produced a larger proportion of false positives, possibly associated with the probabilistic nature of the language model's decision when interpreting report content, in contrast with the deterministic filtering by textual markers employed by the RPA. Even so, the RPA also showed a small number of false positives over the two collection rounds. Investigation of the execution logs identified their cause: the iframe-based extraction strategy, which in some cases selected an incorrect or still-loading frame (Section 4.1). After the frame selection logic was corrected, no new false positives attributable to this cause were recorded.

The exact equivalence between the two solutions in the success rates per step is noteworthy. Both reached a 100\% login rate, a 100\% identification/search rate and a 100\% complete download rate (considering the patients registered in the laboratory). This indicates that, as to the ability to correctly complete the operational flow in the same portal, RPA and APA perform equivalently. The differences between the approaches therefore concentrate on computational cost and execution time, not on execution reliability. Of the patients processed, four could not be located in the laboratory portal by either solution. Later investigation found that they had no active registration in the laboratory system, a limitation of the data environment rather than a failure attributable to the RPA or APA. Both solutions showed exactly the same failure rate in these cases, reinforcing the comparability of the results. Considering only the patients effectively registered in the laboratory (n = 65), login, identification/search and complete download rates reached 100\% for both solutions.

The theoretical advantage of autonomous agents pointed out by \citet{prucha2025}, related to greater adaptability to dynamic interfaces, could not be fully evaluated in this study, since both solutions were evaluated on the same portal, which showed no structural layout changes during the collection period. The results thus reflect the performance of the solutions in a stable-interface scenario and should not necessarily be generalized to contexts of greater interface heterogeneity or instability, in which the flexibility of the APA could represent an advantage not captured by the tests carried out.

As to temporal stability (Section 3.6.4), the comparison between the two collection rounds showed distinct behavior. The RPA remained stable, with variation below 1.5 percentage points in all classification metrics (accuracy: 0.44 p.p.; precision: 1.31 p.p.; recall: 0.00 p.p.; F1-score: 0.68 p.p.). The APA showed substantially larger variation (accuracy: 3.51 p.p.; precision: 4.26 p.p.; recall: 7.04 p.p.; F1-score: 5.59 p.p.). The largest variation occurred in recall, which was 92.96\% at T, when the APA failed to collect five authorized exams, and 100\% at T+7, when all authorized exams were collected. Although the APA performed better in the second round, the variation between repeated executions, on the same patients and reference criteria, indicates lower predictability than that of the RPA. This result diverges from the initial expectation that the flexibility of the autonomous agent would favor greater consistency, and the investigation of the factors responsible for this variation remains a topic for future work.

\subsection{Practical implications}

The operational performance data reveal the main trade-off identified in this study. The mean execution time per patient of the APA (207.21 s) was approximately 13 times that of the RPA (16.07 s), and the cost per report of the APA (USD 0.187) was about 6.2 times higher than that of the RPA (USD 0.030). These results are compatible with the evidence reported in the literature: \citet{huangjoung2026} observed that autonomous agents tend to demand greater computational capacity and present higher operational costs than traditional RPA solutions, which this study also shows quantitatively in a real healthcare application scenario.

For practitioners, the findings suggest that, where the process runs on a single portal with a relatively stable interface, RPA offers lower execution time and cost with equivalent reliability and greater predictability across repeated executions. The adoption of autonomous agents in healthcare automation should consider the characteristics of the process and of the execution environment, and is potentially more pertinent in scenarios that demand greater adaptability to dynamic and heterogeneous interfaces, a condition not fully tested here. The APA design also offers a practical lesson: concentrating the critical steps of filtering, extraction and duplication control in deterministic tools, and keeping the prompt lean, reduces the ambiguity and token consumption of the agent, while the use of containers allows both approaches to be run under equivalent and scalable infrastructure conditions.

\subsection{Limitations and future research}

The study has limitations that suggest directions for future work: (i) the decomposition of the APA cost between the LLM API calls and the cloud infrastructure components, to identify the individual contribution of each factor to the observed total cost; (ii) the evaluation of the solutions in scenarios with greater interface variation, allowing a more direct test of the hypothesis of the APA's greater adaptability; and (iii) the comparison of different language models as the basis of the autonomous agent, to investigate whether the observed differences in cost and performance are inherent to the approach or specific to the model used (Gemini 2.5 Flash-lite). Additionally, the study is limited to a single laboratory portal and to a stable interface during the collection period, and the temporal stability analysis (T vs. T+7) covers only two rounds. The factors that contributed to the variation observed in the APA, especially in recall, which went from 92.96\% at T to 100\% at T+7, deserve deeper investigation.

\section{Conclusion}

This study aimed to compare, systematically and experimentally, the performance of RPA- and autonomous-agent-based (APA) solutions in the automation of a real healthcare process: the extraction of mammography and breast ultrasound reports, contributing to fill the gap identified in the literature regarding the absence of comparative studies between these approaches in heterogeneous clinical environments.

The results indicate that, in a stable-interface scenario, the two solutions are functionally equivalent in execution reliability, both having reached the same success rates in the login, identification/search and complete download steps, but diverge significantly in operational efficiency. The RPA proved to be approximately 13 times faster and 6 times more economical than the APA, besides showing superior classification metrics (accuracy, precision, recall and F1-score). Thus, for the evaluated scenario, a single portal with a relatively stable interface throughout the collection period, the results show lower execution time and operational cost for the RPA, while the APA showed greater variability between executions.

As main contributions, this work offers quantitative evidence on the trade-off between computational cost and flexibility of the two approaches, applied to a real and sensitive healthcare use case, providing an initial reference to support architectural decisions in similar clinical automation projects.

\section*{Declarations}

\bmhead{Author Contributions}

Daniel Schulz: visualization, software development, methodology, investigation and writing -- original draft. Pedro Cabral Buchaim: visualization, software development, methodology, investigation and curation. Bernardo Soares: visualization, software development and investigation. Fernando Rafael Ensslin Schwengber: visualization, investigation, conceptualization, methodology and writing. Lorenzo Farias: software development. Daniela Duarte da Silva Bagatini: conceptualization, supervision, methodology and writing -- review \& editing. Ana Paula Wernz da Cunha Müller: validation and resources. All authors approved the final manuscript.

\bmhead{Funding}

This work was developed within the scope of the project ``Cognitive solution for identifying findings of examinations in the breast cancer care pathway: screening and pre-diagnosis'', affiliated with the Universidade de Santa Cruz do Sul (UNISC). This work was supported by the Fundação de Amparo à Pesquisa do Estado do Rio Grande do Sul (FAPERGS) through the Institutional Scholarship Program for Scientific Initiation and Technological Initiation and Innovation (PROBIC/PROBITI), FAPERGS Call 04/2026, process PROA no. 26/2551-9000039-4.

\bmhead{Data Availability}

The exam reports analyzed during the current study contain sensitive personal health data and, in accordance with the Brazilian General Data Protection Law (LGPD, Law No. 13,709/2018) and the terms of the Informed Consent Form, are not publicly available. The aggregated results supporting the findings are reported within the article. The source code of the APA solution (\url{https://github.com/besoaresn/ClickPalmIA}) and of the RPA solution (\url{https://github.com/Tinpack/RPA_metrificado}) is publicly available. The aggregated per-exam evaluation data, without personal identifiers, are available from the corresponding author on reasonable request.

\bmhead{Ethics Approval and Consent to Participate}

This study and the collection of the reports comply with the Brazilian General Data Protection Law (LGPD, Law No. 13,709/2018). The exams were obtained through the laboratory platform, from patients of a mastologist physician, with approval of the Research Ethics Committee (CEP) through Plataforma Brasil (CAAE 88036825.9.0000.5343). Only mammography and breast ultrasound reports from 2024 to 2025 were included, from patients who authorized the use of their data by signing the Informed Consent Form (TCLE). Pre-operative localization letters and duplicate reports were excluded. The reports were treated exclusively as informational artifacts, without any automated clinical inference, and collection was limited to the minimum data necessary for the study's objectives. In the agent-based solution, the native telemetry of the library used was disabled, avoiding the sending of execution metadata to third parties.

\bmhead{Consent for Publication}

Not applicable. The article contains no individual patient data or identifiable information.

\bmhead{Declaration of Generative AI and AI-Assisted Technologies in the Manuscript Preparation Process}

The authors declare that generative AI tools were used in a limited and supervised capacity during the preparation of this manuscript. Specifically, Claude (Anthropic) was used for the translation of the manuscript from Portuguese into English and for related language editing. The research, its design, analyses, results and scientific content are of human authorship. After each AI-assisted revision, the authors thoroughly reviewed and approved the modified text to ensure its accuracy, coherence, and compliance with the study's objectives.

\bmhead{Competing Interests}

The authors declare no competing interests. None of the authors has any affiliation with the ClickPalm platform. The mastologist physician, whose clinic provided access to the examination reports and collaborated in the case study, is a co-author of this article (Ana Paula Wernz da Cunha Müller); her contributions are described in the Author Contributions section.

\bibliography{refs}

\section*{Appendix A: Online Resources}

\begin{itemize}
\item Online Resource 1: repository of the AI-agent (APA) automation solution: \url{https://github.com/besoaresn/ClickPalmIA}
\item Online Resource 2: repository of the RPA automation solution: \url{https://github.com/Tinpack/RPA_metrificado}
\item Online Resource 3: modified browser-use library: \url{https://drive.google.com/drive/folders/1eq3o6FRCmO3HDvY0Fb-4vrNRo_0c7zUd?usp=sharing}
\end{itemize}

\end{document}